\documentclass[runningheads]{llncs}
\usepackage{graphicx}
\usepackage{amsmath}
\usepackage{booktabs}

\graphicspath{{images/}}
\newcommand{\repeatthanks}{\textsuperscript{\thefootnote}}
\begin{document}
\title{LDRNet: Enabling Real-time Document Localization on Mobile Devices\thanks{ECML-PKDD 2022}}
%
%
\author{Han Wu\inst{1}\thanks{Equal contribution}\and
Holland Qian\inst{2}\repeatthanks\and
Huaming Wu\inst{3}\thanks{corresponding author}\and Aad van Moorsel \inst{4}} 
\authorrunning{H. Wu et al.}
%
\institute{Newcastle University, Newcastle upon Tyne, United Kingdom \\
\email{han.wu@ncl.ac.uk, aad.vanmoorsel@ncl.ac.uk}
\and
Tencent, Shenzhen, China \\
\email{niuwa.frog@gmail.com}
\and
Tianjin University, Tianjin, China\\
\email{whming@tju.edu.cn}
\and
University of Birmingham, Birmingham, United Kingdom \\
\email{a.vanmoorsel@bham.ac.uk}}
\maketitle              
\begin{abstract}
Modern online services often require mobile devices to convert paper-based information into its digital counterpart, e.g, passport, ownership documents, etc. This process relies on Document Localization (DL) technology to detect the outline of a document within a photograph. In recent years, increased demand for real-time DL in live video has emerged, especially in financial services. However, existing machine-learning approaches to DL cannot be easily applied due to the large size of the underlying models and the associated long inference time. In this paper, we propose a lightweight DL model, LDRNet, to localize documents in real-time video captured on mobile devices. On the basis of a lightweight backbone neural network, we design three prediction branches for LDRNet: (1) corner points prediction; (2) line borders prediction and (3) document classification. To improve the accuracy, we design novel supplementary targets, the equal-division points, and use a new loss function named Line Loss. We compare the performance of LDRNet with other popular approaches on localization for general documents in a number of datasets. The experimental results show that LDRNet takes significantly less inference time, while still achieving comparable accuracy.

\keywords{Document localization \and Real time \and Mobile Devices.}
\end{abstract}
\section{Introduction}

The integration of paper documents and digital information is an essential procedure in many online services today. An increasing number of users start to use mobile devices (i.e., smartphones) to take photos of the paper documents. The preliminary step to extract digital information from those photos is Document Localization (DL)~\cite{ref_article1}. DL is a machine learning technology that focuses on detecting and segmenting document outlines within image frames. The input is usually a digital photo containing the paper document and the outputs are the predicted quadrilateral (i.e., four-sided polygon) coordinates of the document outline. Accurate DL is crucial for the follow-up process such as Optical Character Recognition (OCR). 

\begin{figure}[htp]
  \centering
  \vspace{-0.25in}
  \includegraphics[width=0.7\columnwidth]{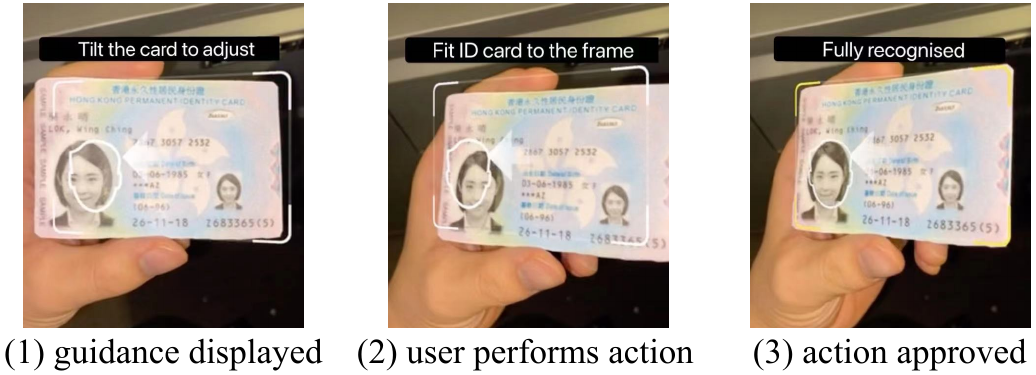}
  \caption{An example of document localization based on video.} \label{fig.frontend}
  \vspace{-0.2in}
\end{figure}

In most online services that use DL, photos captured by mobile devices are uploaded to servers for DL processing. Recently, some service providers, for safety purposes, have started to demand users to capture a video of the paper document instead of a static photo \cite{elecID}. This is because a video is naturally more difficult to counterfeit than a static photo. One concrete example is illustrated in Fig.~\ref{fig.frontend}, where the user uses its smartphone to record a video of the identity document. During the video recording, the mobile application (developed by the service provider) requests the user to move the document properly to fit the guidance displayed on the screen (the white borders in the figures). In the previous design using a static photo, an impostor can cheat the system with a scanned copy of the document. However, in this scheme with a live video it needs to hold the actual document to finish the process. Furthermore, the laser security marks on identity documents change dynamically in the recorded video depending on the light environment and camera angle, which provides more comprehensive materials for the verification process. 

The premise to achieve the above video-based process is that the document outline and trajectory can be tracked in real-time during the video recording. A video is actually a series of images, called frames, that are captured at certain frequency, e.g., 30 Frames Per Second (FPS). Thus DL performed on a video can be understood as a series of DL tasks, where each task is performed on one frame.  
Therefore, real-time DL on a live video means the DL process on each frame needs to be finished within the time interval between two consecutive frames (e.g., 33.3 ms for a 30 FPS video). However, existing DL approaches cannot fulfill these real-time demands due to the long inference time (e.g., over 100 ms even on a PC according to \cite{ref_article18}). Furthermore, state-of-the-art DL models are complex and require large storage space, which potentially exhausts the capacity of mobile devices \cite{ref_article1,das2020hu}.

To break through this bottleneck we propose a novel document localization neural network, \textbf{LDRNet}, to \textbf{L}ocalize \textbf{D}ocument in \textbf{R}eal-time. Previous works dive into the design of the new network architectures to improve the accuracy, which is time-consuming and diminishes the efficiency. We start from a lightweight Convolutional Neural Network (CNN), MobilenetV2~\cite{ref_article22}, which is a fundamental feature extractor especially designed for devices with limited memory and resources. Unlike feature pyramid networks~\cite{ref_article23}, we design a feature fusion module that does not enlarge the model size. Existing DL approaches require postprocessing after prediction, which is cumbersome and inefficient. Therefore we design our prediction target to be the coordinates of the quadrilateral corners instead of the contour of the document thus avoiding postprocessing. The orientation of the document also can be obtained from the order of the output coordinates. We propose a novel loss function, Line Loss, to improve the precision. By adding equal-division points between contiguous corner points, LDRNet achieves better formalization of the borderlines. 

In summary, the main contributions of this paper include:

\begin{itemize}
    \item We present LDRNet, a document localization approach with significantly lower computational cost than the state-of-the-art methods. LDRNet paves the way for real-time DL on a live video recorded by mobile devices. 
    \item We design the Line Loss function and equal-division points feature for LDRNet to guarantee the localization accuracy without undermining its efficiency or enlarging its model size.
    \item In the experiments, we compare the performance of LDRNet with other popular DL approaches on localizing general document datasets. The results indicate that LDRNet achieves comparable accuracy while outperforming other approaches in terms of efficiency. 
\end{itemize}

\section{Related Work}
There exist three main kinds of approaches for DL: Mathematical Morphology-based Methods, Segmentation-based Methods and Keypoint-like Methods. Mathematical morphology-based methods are based on mathematical morphology~\cite{ref_article9}. There are some other hand-designed features used in mathematical morphology-based methods, like the tree-based representation~\cite{ref_article11}. Along with the popularity of CNN in this field, many CNN-based methods have emerged. Segmentation-based methods regard DL as the segmentation~\cite{ref_article17} task using the CNN to extract the features. Same as segmentation-based methods, using the features extracted by the CNN, keypoint-like methods~\cite{ref_article18} predict the four corners of the document directly, considering DL as the keypoint detection task.

\textbf{Mathematical Morphology-based Methods} inherit the ideas which detect the contour of the documents using traditional image processing methods, image gradients calculations~\cite{ref_article9}, Canny border detectors, Line Segment detectors~\cite{ref_article10} and image contours detectors, etc. Although there are many kinds of different mathematical morphology-based approaches, they are all developed on the basis of the technologies mentioned above, which makes the performance unstable when the datasets change. The accuracy of these methods heavily depends on the environmental conditions in the image. For instance, if the color of the background and the document are difficult to distinguish, or if the image is captured with insufficient lighting, the borders of the document may not be detected. Another weakness of these mathematical morphology-based methods is that they output the four borders or four points disorderly so a necessary step for determining the orientation of the document is the postprocessing, which leads to extra cost.

\textbf{Segmentation-based Methods} regard DL as a segmentation task. Segmentation adapts dense predictions, outputs the heat map for every pixel on the image, and uses classification labels to determine whether the pixels belong to the object or the background. Then by grouping the pixels with the same labels, the document is segmented. By adopting the CNNs to extract the image feature, the segmentors get rid of the impacts from the complex environment conditions. Since every segmentor is a data-driven deep-learning model, it can reach high precision as long as enough data are fed. U-Net~\cite{ref_article23} and DeepLab~\cite{ref_article24} are the popular segmentors. However, the large model size and long inference time make these segmentors incompetent for real-time DL. Similar to the mathematical morphology-based methods, postprocessing is inevitable to find the orientation of the document content.

\textbf{Keypoint-like Methods} output the coordinates of the quadrilateral corner points of the document directly. Recent keypoint detection networks do not regress the coordinates of the key points, instead, they produce dense predictions like segmentation networks do. \cite{ref_article28} predict heat maps of the keypoints and offsets. \cite{ref_article18} predict the points in a sparse-prediction way to locate the four points directly. To improve the precision, it uses CNN recursively to fix the coordinates errors. These key-point models indeed get high precision, but also have the same weakness which segmentation-based methods have, the large model size and the long inference time.

\section{Context and Methodology}

\subsection{Problems and Challenges}

In previous online services, DL task is performed on the server while the mobile device only captures and uploads the photo of the document. This structure can not fulfil the real-time DL task on a video due to the transmission cost. Therefore we aim to embed DL module on mobile devices in our work. 
Tracking the document outline and trajectory in a live video means the DL process for each frame should be completed within the frame interval (33.3 millisecond for a 30 FPS video). This calls for strict demands on both the accuracy and speed of DL model. 

Specifically, the challenges of this study come from four facets: (i) The computational resource on mobile devices is very limited while existing DL approaches require large memory and long inference time. (ii) In addition to the contour of the document, the direction of the content should also be detected to determine the trajectory of the document in a video. (iii) It is complex and time-consuming to calculate the precise angle between the document and the camera to obtain the trajectory. (iv) During the video recording, the corner points of the document may be occluded by the user's fingers, therefore the ability to predict occluded corner points is necessary.

\subsection{Task Analysis} 

To address the challenges listed above, we present a novel neural network model, LDRNet, to Localize Documents in Real-time. Instead of calculating the precise angle between the document and camera, we calculate the distance between each target corner point and the corresponding localized point to track the trajectory of the document. This provides considerable accuracy while consuming less computational resources on mobile devices. As summarized by the following equation, $(x_{doc}^i,y_{doc}^i)$ is the coordinate of the $i$th corner point of the localized document, while $(x_{target}^i,y_{target}^i)$ represents the coordinate of the $i$th target corner point. Then we sum the Euclidean distances of the four sets of corresponding points.
\begin{equation}
    Distance = \sum_{i=1}^{4} \sqrt{(x_{doc}^i-x_{target}^i)^2 + (y_{doc}^i-y_{target}^i)^2}.
     \label{equ.fe_dist} 
 \end{equation}

The orientation of the document can be simply inferred from the order of the corner points. Thus our goal is to predict the four quadrilateral coordinates of the document in counter-clockwise order. The order of the four quadrilateral points is determined by the contents of the document instead of the direction that the document is placed. Throughout this paper, we use $N$ to denote the total number of points we predict for each document. In addition to the four corner points, we predict $(N-4)/4$ equal-division points on each border of the document. These extra $N-4$ points are used to refine the localization of the document. Moreover, we add a classification head to our network architecture for classifying the document in the input images. Depending on the specific DL task, this classification head is adjustable. The minimum number of classes is two, which represents whether the image contains a document or not, respectively.

\subsection{Network Architecture}

\begin{figure}[htp!]
  \centering
  \includegraphics[width=0.95\textwidth]{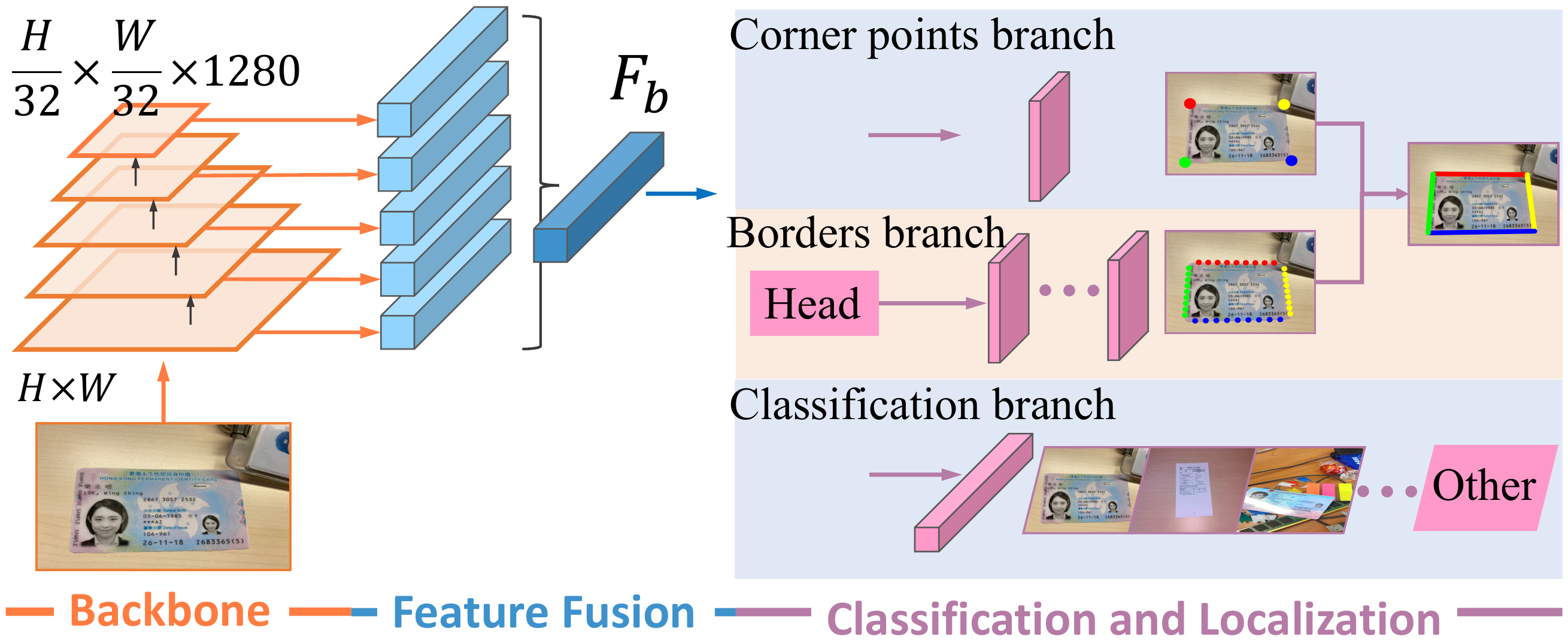}
  \caption{The network architecture of LDRNet.} \label{fig.2}
\end{figure}
 
\textbf{Fully Convolutional Feature Extractor.} As we aim to run DL on mobile devices, we choose a lightweight backbone network, MobilenetV2~\cite{ref_article22}. It applies both depth-wise convolution and point-wise convolution operations to achieve faster and lighter extraction. As illustrated in Fig.~\ref{fig.2}, the last output feature map from the backbone is $F_b\in R^{\frac{H}{32}\times\frac{W}{32}\times1280}$ with $H$ denoting the height of the input image and $W$ denoting the width. To improve the accuracy, we extract five feature maps with different spatial resolutions from the backbone.

\textbf{Feature Fusion Module.} 
The low and high-level feature maps are fused together by the feature fusion module. The first step is feature compression, where we use global average pooling to downsample the feature maps, and resize them to the same size. Then we add the five feature maps directly instead of the top-down architecture used in \cite{ref_article25}. 

\textbf{Network Output Branches.} The outputs of the LDRNet consist of three branches. The first branch is the corner points branch. It outputs in the form of a
$4\times 2$ vector, four corners' coordinates $(x,y)$ in order. The second branch is the borders branch, it outputs in the form of an $(N-4)\times 2$ vector, where $(N-4)$ is the number of points to be predicted on the four borders. Each border has $(N-4)/4$ points so there are $N-4$ coordinates of points in total on the second branch. The third branch outputs the classification label, denoting the type of document in the input image. Unless the size of the classification output is specified, the classification output contains two elements, one denoting the likelihood of having documents in the image, the other one denoting the likelihood that no document is detected in the input image. 

\textbf{Line Loss.} Standard Deep Convolutional Neural Network architectures are inherently poor at precise localization and segmentation tasks~\cite{ref_article30}. This is because the last convolutional layer only contains 
high-level features of the whole image. While these features are extremely useful for classification and bounding box detection, they lack the information for pixel-level segmentation~\cite{ref_article18}. In order to improve the
precision of DL, we combine the two branches of the LDRNet's outputs (corner points branch and borders branch), we predict the corners in a line-prediction fashion. In addition to the four corner points, we also predict the equal-division points on the lines thus the labels can be generated automatically and no more human effort is required. The proposed \textbf{Line Loss} is formulated as $L_{line}(p) =\beta L_{Sim}(p) + \gamma L_{Dis}(p)$, which is a weighted sum of the similarity loss $L_{Sim}$ and the distance loss $L_{Dis}$. The similarity loss is used to restrict the points from the same border along an identical line, while the distance loss is used to guarantee that along this line the points are equally divided.

To guarantee that the predicted points from each border are on a straight line, we use the \textbf{similarity loss} $L_{Sim}$ to calculate the similarity of two vectors of the three successive points on the line. The details of $L_{Sim}$ are shown in Eq.~(\ref{equ.sim1}), (\ref{equ.sim2}), (\ref{equ.sim3}).
\begin{align}
   & L_{Sim}(p) =[\sum_{k\in{l,r,t,b}} \sum_{i=0}^{\frac{N}{4}-3} sim(p[k]_i,p[k]_{i+1},p[k]_{i+2})]/(N-4), 
    \label{equ.sim1} \\
  &  sim(p[k]_i,p[k]_{i+1},p[k]_{i+2}) = (\overrightarrow{p[k]}_i^{i+1} \cdot \overrightarrow{p[k]}_{i+1}^{i+2})/ (\left| \overrightarrow{p[k]}_i^{i+1} \right| \times \left| \overrightarrow{p[k]}_{i+1}^{i+2} \right| ), 
    \label{equ.sim2} \\
  &  \overrightarrow{p[k]}_i^{i+1} = \left( p[k]_i^x - p[k]_{i+1}^x,p[k]_i^y - p[k]_{i+1}^y \right). 
    \label{equ.sim3} 
\end{align}
where $p[l]$, $p[r]$, $p[t]$, $p[b]$ denote the points on the left border, on the right border, on the top border and on the bottom border, respectively.

The \textbf{distance loss} is used to constrain the points we predict to be equal-division points. We use Eqs.~(\ref{equ.dis1}) and (\ref{equ.dis2}) to make sure the successive points of each border have the same distance in both x-direction and y-direction. 
\begin{align}
      &  L_{Dis}(p) = [\sum_{k\in{l,r,t,d}} \sum_{i=0}^{\frac{N}{4}-1} dist(p[k]_i,p[k]_{i+1},p[k]_{i+2})]/(N-4),
        \label{equ.dis1} \\
        &dist(p[k]_i,p[k]_{i+1},p[k]_{i+2}) = \left| \left|p[k]_i^x-p[k]_{i+1}^x\right| -\left|p[k]_{i+1}^x-p[k]_{i+2}^x\right| \right| + \nonumber\\
        & \left| \left|p[k]_i^y-p[k]_{i+1}^y\right| -\left|p[k]_{i+1}^y-p[k]_{i+2}^y\right| \right|.
        \label{equ.dis2} 
\end{align}

Furthermore, we use L2 loss for the regression and cross-entropy for the classification. The \textbf{regression loss} $L_{Reg}$ is an L2 loss between the predicted points $p$ and the ground truth points $g$, which can be formulated as:
\begin{align}
    L_{Reg}(p,g) &= \frac{1}{N-4} \sum_{i=0}^{N} \sum_{j\in{x,y}} \sqrt[2]{(\hat{g}_{i}^{j}-p_{i}^{j})^2}, &&
    (\hat{g}^x = g^x/W,\hat{g}^y = g^y/H).
    \label{equ.reg1}
\end{align}
where $(g_i^x, g_i^y)$ denotes the $i$-th ground truth point of the document. Our regression target is $\hat{g}$, which is the normalization of $g$ by image width ($W$) in x-coordinate and image height ($H$) in y-coordinate.

The \textbf{classification loss} $L_{Cls}$ is soft-max loss over multiple classes confidences (x), which is calculated as:
\begin{align}
    L_{Cls}(x,c)& = \sum_{i=0}^{N_{cls}} -c_i \log \hat{x}_i,& (\hat{x}_i = \frac{\exp(x_i)}{\sum_j \exp(x_j)}).
    \label{equ.cls1}  
\end{align}
where $c_i \in \{0,1\}$ is an indicator denoting whether the
image contains the $i$-th category document and $N_{cls}$ is the number of the total document categories.

Finally, we define the total loss as the weighted sum of the regression loss $L_{Reg}$, the classification loss $L_{Cls}$ and the Line Loss $L_{Line}$:
\begin{align}
    L(x,c,p,g) &= L_{Reg}(p,g) + \delta L_{Cls}(x,c) + L_{line}(p).
\end{align}
where the weights $\delta$, $\beta$ and $\gamma$ are chosen depending on the experimental results, and the values normally range from 0 to 1.

\section{Experimental Evaluation}

For the comparison experiment, we use the dataset from `ICDAR 2015 SmartDoc Challenge 1'~\cite{ref_article1}. Training and inference setting details are listed in this section. The experimental results are compared to the previous work to show the advantages of our approach. Then we use the ablation study to analyze the contribution of each component of our model. Finally, we test our model on the MIDV-2019 dataset \cite{bulatov2020midv} to highlight the characteristic of our model, the ability to predict occluded corner points. 

\subsection{Training and Inference Details}

Unless specified, we use MobilenetV2 with the width multiplier $\alpha$ equal to 0.35 (used to control the width of the network) as our backbone network. We set the number of regression points ($N$) to 100. Our network is trained with RMSprop optimizer, which uses only one set of hyperparameters (rho is set to 0.9, momentum is set to 0, and epsilon is set to 1e-7). We trained our networks for 1000 epochs, with an initial learning rate of 0.001 and a batch size of 128 images. The learning rate is reduced in a piecewise constant decay way, and is set as 0.0001, 0.00005, 0.00001 at the 250th, 700th and 850 epochs, respectively. Our backbone network weights are initialized with the weights pretrained on ImageNet~\cite{ref_article31}. We use the Xavier initializer~\cite{ref_article32} as the final dense layer. The input images are resized to which both the width and the height are 224 pixels. Regarding the Line Loss function parameters, $\delta$ is set to 0.32, $\beta$ and $\gamma$ are configured as 0.0032.

For the inference, we first forward the input image through the network to obtain the quadrilateral points' coordinates of the documents and the predicted class. Then we multiply the quadrilateral points' coordinates by the width ($W$) and height ($H$) of the input image.
Note that we only use four quadrilateral points' coordinates instead of the predicted $N$
coordinates, because we found little difference between their performance. Thus we can
remove the weights of the final dense layer that are not used for the four quadrilateral coordinates. The size of the input image is the same as we used for training.

\subsection{Comparison of Accuracy}

To evaluate the accuracy of our DL model, we use the Jaccard Index (JI), which is also adopted in others' work \cite{ref_article1,ref_article18,das2020hu}. First we remove the perspective transform of the ground-truth $G$ and the predicted results $S$, then obtain the corrected quadrilaterals $S^{'}$ and $G^{'}$. For each frame $f$, the JI is computed as $JI(f) = area(G^{'} \cap S^{'})/area(G^{'} \cup S^{'})$. The value of JI range from 0 to 1 and higher JI indicates higher accuracy.

\begin{table}[]
  \centering
  \caption{Accuracy compared with previous works. The results are listed from top to bottom in the descending order of overall JI.}\label{tab1}
  \begin{tabular}{p{3.2cm}p{1.3cm}p{1.3cm}p{1.3cm}p{1.3cm}p{1.3cm}p{1.3cm}}
  \toprule
  \textbf{Method}& \multicolumn{5}{l}{\textbf{Background}} & \textbf{Overall} \\ \cline{2-6}
  \rule{0pt}{2ex} & 01   & 02   & 03   & 04  & 05  &         \\ \hline    HU-PageScan~\cite{das2020hu} &  / & /&/&/&/&\textbf{0.9923}\\
  \textbf{LDRNet-1.4 (ours)}            &        0.9877          &  \textbf{0.9838}    &  0.9862   &      0.9802     &  \textbf{0.9858}   &    0.9849   \\
  SEECS-NUST-2~\cite{ref_article18} &  0.9832 & 0.9724&0.9830&0.9695&0.9478&0.9743\\
  LRDE~\cite{ref_article1}          &  0.9869 & 0.9775&0.9889&\textbf{0.9837}&0.8613&0.9716\\
    SmartEngines~\cite{ref_article1}  &  \textbf{0.9885} & 0.9833&\textbf{0.9897}&0.9785&0.6884&0.9548 \\
  NetEase~\cite{ref_article1}      &   0.9624 & 0.9552&0.9621& 0.9511&0.2218&0.8820 \\
  RPPDI-UPE~\cite{ref_article1}     &  0.8274 & 0.9104&0.9697& 0.3649&0.2163&0.7408 \\
  SEECS-NUST~\cite{ref_article1}    &  0.8875 & 0.8264&0.7832&0.7811&0.0113&0.7393\\

 \bottomrule
  \end{tabular}
  \end{table}

As shown in Table~\ref{tab1}, the images in the dataset can be divided into five categories according to different backgrounds. Only backgound05 is complex, with strong occlusions. We compare the accuracy of LDRNet to seven previous DL models. It is observed that our LDRNet outperforms the previous works in terms of background02 and background05 (results in bold). For other backgrounds, LDRNet reaches comparable performance with the best ones. The overall JI of LDRNet exceeds the other methods except for HU-PageScan in \cite{das2020hu}, which does not provide the results of background01 to background05. However, HU-PageScan uses 8,873,889 trainable parameters which is over 21 times the number of parameters in our LDRNet-0.35 (denotes LDRNet with $\alpha=0.35$). Therefore HU-PageScan requires significant memory and computing time thus can not fulfill the real-time demand. This will be introduced in the next section. Additionally, since HU-PageScan is segmentation-based, it only predicts the contour of the document. Thus the orientation of the document is unknown and requires follow-up process to calculate the document trajectory.

\subsection{Comparison of Inference Time}

\begin{figure}[ht]
  \centering
  \vspace{-0.3in}
  \includegraphics[width=0.87\textwidth]{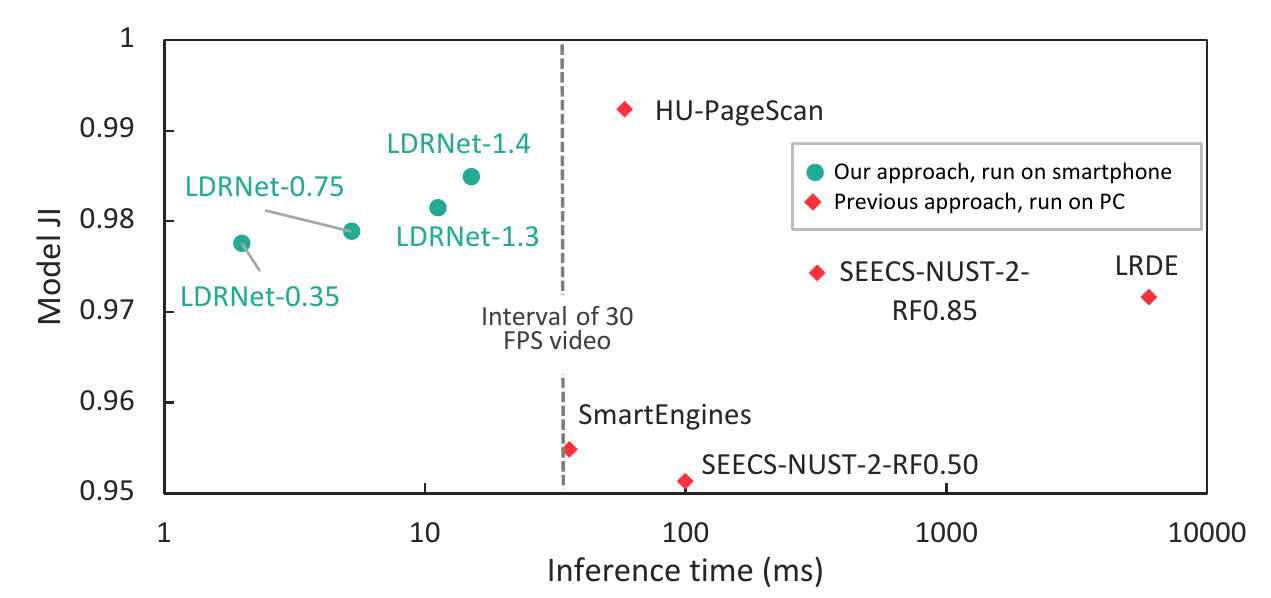}
\caption{The inference time comparison between LDRNet and the previous DL methods on the `ICDAR 2015 SmartDoc Challenge 1' dataset. The horizontal axis is log scaled.}\label{fig.JI_speed}
\end{figure}

Our Network is tested on iPhone11 using TNN engine. HU-PageScan is tested on a PC equipped with Intel Core i7 8700 processor, 8GB RAM, and 6GB NVIDIA GTX 1060~\cite{das2020hu}. In Fig.~\ref{fig.JI_speed}, the vertical axis is the JI of the model while the horizontal axis is the log scaled inference time. We illustrate the result of four settings of LDRNet, all using MobilenetV2 but with different values of $\alpha$ (0.1, 0.35, 0.75, 1.3, 1.4). We observe that higher $\alpha$ leads to higher JI but longer inference time. The JI of HU-PageScan (run on a PC) is 0.0074 (absolute value) higher than LDRNet-1.4 (run on smartphone), whereas the inference time is about 4x longer. The short inference time of LDRNet meets the demand for localizing documents in the image frames in a live video (usually photographed at 30 FPS, represented by the dashed vertical line in Fig.~\ref{fig.JI_speed}). For general usage, LDRNet-1.4 is the best option and its model size is only 10MB.

\subsection{Ablation Study}\label{sec_featureFusionExp}


\begin{figure}[htbp]
  \centering
  \vspace{-0.3in}
  \includegraphics[width=0.75\columnwidth]{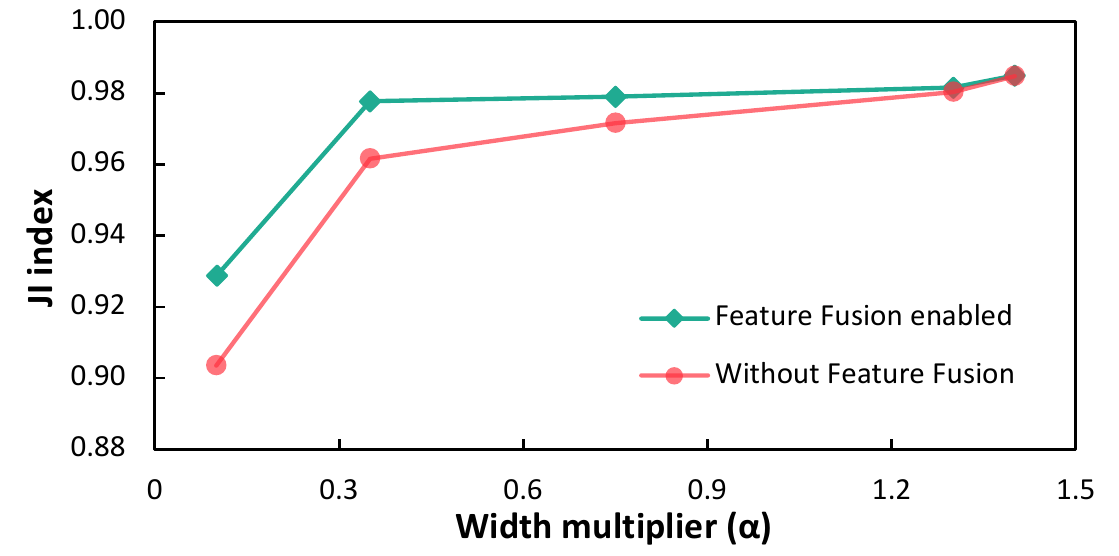}
  \caption{The JI of LDRNet with different $\alpha$ and with or without feature fusion module. The number of regression points is set to 100. All are trained with Line Loss.} 
  \label{fig_ffefficiency}
\end{figure}

In our experiments using LDRNet, we construct the feature fusion module using \textit{average pooling} and \textit{add} operation. To evaluate the efficiency of this feature fusion module, we run experiments with this module enabled and disabled. Fig.~\ref{fig_ffefficiency} compares the JI of these two scenarios with $\alpha$ ranging from 0.1 to 1.4. We can observe that the feature fusion-enabled models outperform those without feature fusion. Since the model complexity grows as we increase $\alpha$, it is observed that the efficiency of our feature fusion module drops as the model becomes more complex. Thus in the cases that $\alpha>1.0$, feature fusion is not recommended.

We also evaluate the efficiency of the Line Loss by comparing the JI of models with and without Line Loss. For LDRNet-0.35, enabling Line Loss improves the JI from 0.9643 to 0.9776.

\subsection{Predictions of the Occluded Points}

\begin{figure}[htp]
  \centering
   \vspace{-0.1in}
  \includegraphics[width=0.8\columnwidth]{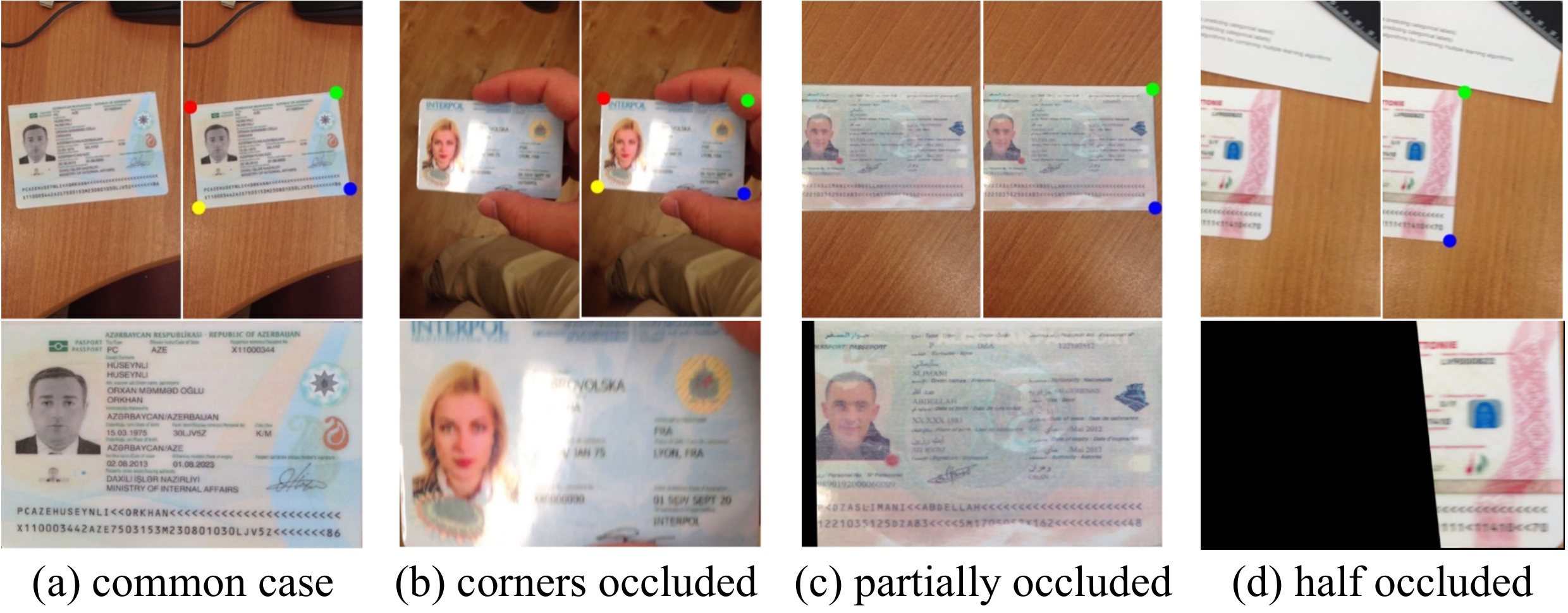}
  \caption{Examples of occluded points prediction. Each case contains three images, namely, the input image (top left), the predicted corners on the input image (top right), the localized document after removing the perspective transformation (bottom). } 
  \label{fig.midv}
  \vspace{-0.15in}
\end{figure}

Benefiting from the task analysis and the network architecture, LDRNet is able to predict the occluded points, including the points occluded by other objects and the points out of the input image. This characteristic is crucial for video recording since the document is usually occluded by the user's fingers during the interaction. For evaluation we test our model on the  MIDV-2019 dataset, which contains video clips of identity documents captured by smartphones in low light conditions and with higher projective distortions \cite{bulatov2020midv}. As depicted in Fig.~\ref{fig.midv}(b), LDRNet can predict the corner occluded by fingers. Even if more than half of the passport is out of the image, as illustrated in Fig.~\ref{fig.midv}(d), our LDRNet predicts the occluded corners correctly.

\section{Conclusion}

We design LDRNet, a real-time document localization model for mobile devices. LDRNet extracts the image features using neural networks and predicts the coordinates of quadrilateral points directly. We propose the novel loss function, Line Loss, and design the equal-division points feature to guarantee its efficiency and accuracy. The most practical scenario of LDRNet is tracking the trajectory of document in a live video captured on mobile devices. The experimental results show that LDRNet has lower inference time than other methods, while achieving comparable accuracy. Currently, LDRNet is being deployed in the identity verification system of a company that serves about 3.8 million customers. The code is available at: \url{https://github.com/niuwagege/LDRNet}. In future work, we will finetune the hyper-parameters more precisely, use low-level and high-level image features fusions like FPN, or a larger backbone, etc.


%
%
%
%




\bibliographystyle{splncs04}
\bibliography{ldrnet}
\end{document}